\documentclass[sigconf]{acmart}

\usepackage{enumitem}

\acmYear{2021}
\copyrightyear{2021}
\setcopyright{acmlicensed}
\acmConference[GECCO '21]{2021 Genetic and Evolutionary Computation Conference}{July 10--14, 2021}{Lille, France}
\acmBooktitle{2021 Genetic and Evolutionary Computation Conference (GECCO '21), July 10--14, 2021, Lille, France}
\acmISBN{978-1-4503-8350-9/21/07}\acmPrice{15.00}
\acmDOI{10.1145/3449639.3459287}

\begin{document}
\title{Expressivity of Parameterized and Data-driven Representations in Quality Diversity Search}

\author{Alexander Hagg}
\email{alex@haggdesign.de}
\orcid{0000-0002-8668-1796}
\additionalaffiliation{
	\institution{Leiden Institute of Advanced Computer Science}
	\city{Leiden}
	\country{The Netherlands}
}
\affiliation{
  \institution{Bonn-Rhein-Sieg University of Applied Sciences}
  \city{Sankt Augustin}
  \country{Germany}
}

\author{Sebastian Berns}
\orcid{0000-0001-9555-5954}
\affiliation{
  \institution{Queen Mary University of London}
  \city{London}
  \country{United Kingdom}
}

\author{Alexander Asteroth}
\orcid{0000-0003-1133-9424}
\affiliation{
  \institution{Bonn-Rhein-Sieg University of Applied Sciences}
  \city{Sankt Augustin}
  \country{Germany}
}

\author{Simon Colton}
\orcid{0000-0002-4887-6947}
\additionalaffiliation{
	\institution{SensiLab, Monash University}
	\city{Melbourne}
	\country{Australia}
}
\affiliation{
  \institution{Queen Mary University of London}
  \city{London}
  \country{United Kingdom}
}

\author{Thomas B\"ack}
\orcid{0000-0001-6768-1478}
\affiliation{
  \institution{Leiden Institute of Advanced Computer Science}
  \city{Leiden}
  \country{The Netherlands}
}

\renewcommand{\shortauthors}{A. Hagg et al.}

\begin{abstract}
We consider multi-solution optimization and generative models for the generation of diverse artifacts and the discovery of novel solutions.
In cases where the domain's factors of variation are unknown or too complex to encode manually, generative models can provide a learned latent space to approximate these factors.
When used as a search space, however, the range and diversity of possible outputs are limited to the expressivity and generative capabilities of the learned model. 
We compare the output diversity of a quality diversity evolutionary search performed in two different search spaces: 1) a predefined parameterized space and 2) the latent space of a variational autoencoder model.
We find that the search on an explicit parametric encoding creates more diverse artifact sets than searching the latent space. A learned model is better at interpolating between known data points than at extrapolating or expanding towards unseen examples. 
We recommend using a generative model's latent space primarily to measure similarity between artifacts rather than for search and generation.
Whenever a parametric encoding is obtainable, it should be preferred over a learned representation as it produces a higher diversity of solutions.
\end{abstract}



\sloppy
\maketitle


\section{Introduction}

\begin{figure}[htb]
    \centering
    \includegraphics[width=\linewidth]{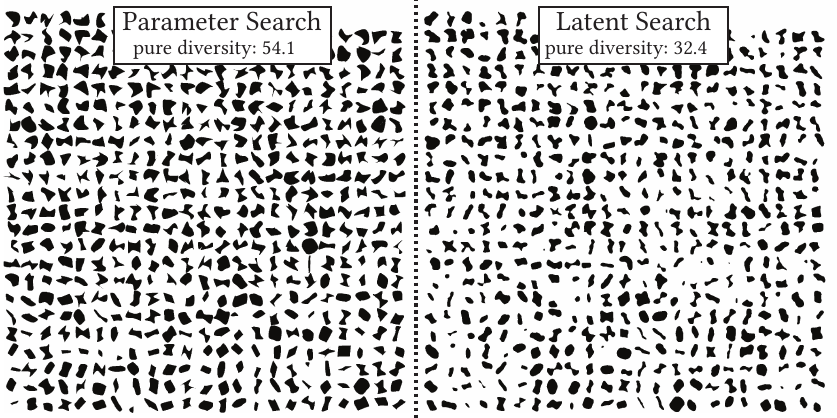}
    \caption{Searching the parameter space produces a more diverse set of artifacts than searching a VAE's latent space. In both cases, the same VAE's latent dimensions were used as niching dimensions of a quality diversity algorithm.}
    \label{fig:QDshapes}
\end{figure}

\noindent
While engineering-driven design optimization looks for solutions to technical problems, artistic practices are usually more concerned with generating culturally valuable artifacts. However, these two approaches are more similar than the seeming difference in focus and objective would suggest. Architects and engineers often use the output of a design optimization tool in the beginning of the design process in order to survey the space of possibilities, where underlying parameters can have complicated correlations~\cite{bradner2014parameters}. Candidate solutions are then expanded or contracted upon in an iterative design loop. Similarly, artists might set up an evolutionary system to find initial inspiration and continue to it towards a desired outcome through the iterative adjustment of the fitness function. 
In both workflows, the diversity of the generated population is key to illustrating the range of possibilities. We propose that initial diversity is the basis for the potential of later discoveries. Focusing on only one optimal individual too early limits the chances of encountering unexpected candidate solutions. 

Evolutionary multi-solution approaches such as quality diversity (QD) algorithms have been developed for the purpose of divergent search~\cite{Lehman2011}. 
Defining QD descriptors by hand is a non-trivial task which requires expertise and, depending on the domain, often cannot compete with an automated solution~\cite{Hagg2020b}.
Deep generative models (GM) such as variational autoencoders (VAE)~\cite{kingma2014autoencoding} can extract patterns from raw data, learn meaningful representations for the data set and accurately produce more samples with similar properties. Disentangled representation learning can furthermore equip a model's latent space with linearly separated factors of variation~\cite{burgess2017understanding}, revealing the underlying factors of a generative process. The resulting feature compression model encodes descriptors to be used with QD algorithms~\cite{cully2019autonomous,gaier2020,Hagg2020b}. While the advantage of learning from data lies in the recognition of complex patterns, the expressivity of the resulting GM is entirely dependent on the quality and representativeness of the data samples provided. This is especially critical when relying on such a model to produce novel examples and diverse sets of outputs. In fact, artists who employ generative adversarial networks (GANs) often use a variety of strategies to actively diverge from the intended purpose of these models and produce outputs significantly different from the original data~\cite{berns2020bridging}.

We compare the performance of multi-solution evolutionary search in the parameter space of a generative system with the search in the latent space of a VAE that was trained with examples from the same system. An example of the resulting solution sets (see Sec.~\ref{experiments:II}) produced by the two search methods is depicted in Fig.~\ref{fig:QDshapes}. While the latent space is built from a limited data set, the parameter space represents the full range of the system's possible output. The purpose of this work is to understand how expressive either of these search spaces are and, from this knowledge, to derive recommendations for their usage.
We choose the simple, yet illustrative example problem of shape optimization, as previously introduced~\cite{Hagg2020b}, for easy interpretation and visualization. 
While more complex domains might be closer to actual applications, they would make presentation of our results less accessible. We assume our findings generalize to those domains.
Shape is an important basic design element in art, architecture, engineering, as well as graphic and industrial design. On the one hand, shapes can carry semantic meaning (e.g. letters of a font) and on the other hand, define the properties and visualize the form of a physical object in engineering-driven design (e.g. the cross-section of a wing optimized for aerodynamical flow). 

Our work is relevant in two scenarios: 1) when the generative process is manually defined but a VAE is used to compare artifacts (i.e. distance/similarity estimation), and 2) when only data is available and the underlying patterns are unknown or too difficult to extract manually and have to be learned by an appropriate model. The present study makes the following contributions: 
\begin{enumerate}
    \item In the context of the first scenario, we give informed recommendations of how to use a VAE to its full capacity in combination with a QD algorithm. We test whether the latent space is suitable both for searching for artifacts and for evaluating artifacts' similarity or whether the two steps should be performed in separate spaces.
    \item For both scenarios, we give evidence for the limitations of VAEs in their ability to represent and generate examples beyond the original training data and, as a result, the diversity of their output.
\end{enumerate}

\section{Background}
In this section, we provide background knowledge on the two core methods used in our generative system, VAE and QD search. We briefly discuss related work.

\subsection{Variational Autoencoders} 

\begin{figure}[tb]
	\centering
	\includegraphics[width=\linewidth]{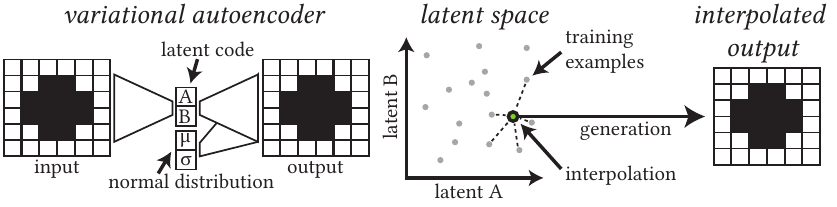}
	\caption{Left: variational autoencoder, center: sampling from latent space, right: interpolated output.}
	\label{img:vae}
\end{figure}

VAEs are a likelihood-based method for generative modelling in deep learning. They follow the standard architecture of an auto\-encoder: a compressing encoder network, mapping data samples to latent space, and a decoder network which is trained to generate the original samples from the corresponding latent codes (Fig.~\ref{img:vae}). A VAE can generate new samples by interpolating between training locations in the latent space. While common autoencoders draw from an unrestricted range of latent code values, the latent space of a VAE is typically modelled to be a centered isotropic multi-variate Gaussian ($\mathcal{N}(0,I)$).
The VAE training objective is to optimize a lower bound on the log-likelihood of the data. We use a beta-annealing variant of the loss term to improve disentanglement with improved reconstruction \cite{burgess2017understanding}. This variant of the evidence lower bound (ELBO) calculates the loss function $L$ over the predicted output $x$ and the ground truth $\hat{x}$ as follows:

\begin{equation}
L(x,\hat{x}) = C(x,\hat{x}) + \beta \cdot (KL(x,0,1) - \gamma)
\label{eq:VAEloss}
\end{equation}

\noindent
Eq.~\ref{eq:VAEloss} consists of a reconstruction loss term, in this case the binary cross-entropy $C$ between prediction and ground truth, and a regularization term, which penalizes a latent distribution that is not similar to a normal distribution with $\mu=0$ and $\sigma=1$. The regularization term is calculated using Kullback-Leibler divergence and scaled by the parameter $\beta$. The annealing factor $\gamma$ is increased from 0 to 5 during training to focus on improving the reconstruction error in the beginning of the training and then gradually improve the distribution in latent space. The internal latent space of a converged model provides meaningful representations in which distances between data points correspond to their semantic similarity. In this work, we use a VAE's internal representation to estimate the similarity of artifacts. 

Previous work employed autoencoders for dimensionality reduction and the encoding of behavioral descriptors in a control task. 
In robotics, this approach allows robots to autonomously discover the range of their capabilities, without prior knowledge~\cite{cully2019autonomous}. GM have been used to distinguish parameterized representations in shape optimization~\cite{Hagg2020d,Hagg2020b}. They have also been employed to learn an encoding during optimization, using them as a variational operator~\cite{gaier2020}. Other GMs like GANs have been used in latent variable evolution~\cite{bontrager2018deepmasterprints} to generate levels for the video games Super Mario Bros.~\cite{volz2018evolving} and Doom~\cite{giacomello2019searching}. A model's latent space is searched with an evolutionary algorithm for instances that optimize for desired properties such as the layout or difficulty of a level. While some authors view the generated levels as novel, none have studied exactly how novel or diverse of an output such a system can produce. 

\subsection{Quality Diversity Search} 
\label{sec:QD}
Optimality is not always the only goal in engineering or design. Finding a diverse set of ways to solve a problem increases potential innovation in the design process. Algorithms built around diversity as well as optimality enable engineers to use algorithms much earlier in the real world design process. Multi-solution optimization is a field that is getting more attention due to the advent of QD algorithms and GM. QD has been shown to produce more diverse sets of artifacts than classical approaches like multi-criterion and multimodal optimization~\cite{Hagg2020b}. 

Multi-criterion optimization defines diversity w.r.t. solution fitness. Multimodal optimization uses parametric similarity to distinguish and protect novel solutions, creating a diverse set of artifacts. 
In contrast, QD compares solutions on the basis of phenotypic, not parametric or objective similarity, and combines optimality with solution diversity~\cite{Hagg2020c}. QD measures similarity between artifacts based on morphological or behavioral features that can usually only be obtained by expressing a solution to its phenotype or even placing the artifact in its environment, i.e. through expensive simulation.

\begin{figure}[tb]
	\centering
	\includegraphics[width=\linewidth]{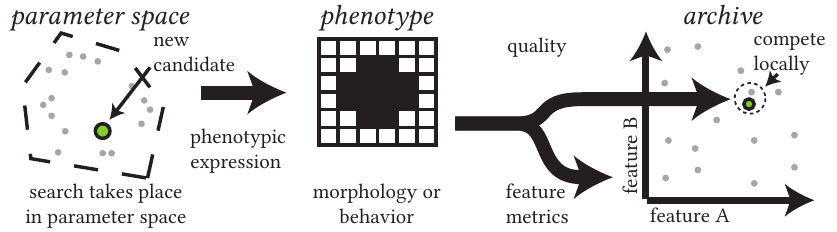}
	\caption{QD searches in parameter space and maintains diversity by only allowing solutions to compete based on their phenotypic similarity. Candidate artifacts are assigned to an archive based on features that are usually manually created a priori. Candidates are only placed inside the archive if they improve its quality value locally.}
	\label{img:archive}
\end{figure}

QD searches in parameter space (Fig.~\ref{img:archive}), but solutions are evaluated based on their expressed phenotypes. Predefined feature metrics, which measure some aspects of behavior or morphology are used to assign an artifact to a niche in an archive, which keeps track of the artifacts found so far. Niching is commonly used in evolutionary approaches to protect novel solutions from not being selected. Examples of archive features are the proportion of time that each leg is in contact with the ground in a hexapod robot's walking gait~\cite{Cully2015}, the turbulence in the air flow around a shape~\cite{Hagg2020e}, or surface area of a shape~\cite{Hagg2020e}. Competition between artifacts only takes place when they are assigned to the same niche. An artifact is only added if it survives local competition in the niche. New artifacts are created by selecting surviving candidates from the archive and adding perturbations to their genome, e.g. through mutation and/or crossover with other genomes. 

The QD algorithm used in this work is based on an alternative formulation to MAP-Elites~\cite{Cully2015}. \textit{Elites} is the common nomenclature for high-performing archive members. In MAP-Elites, the archive consists of a fixed grid of niches, which leads to an exponential growth of niches with the number of phenotypic feature dimensions. CVT-Elites~\cite{Vassiliades} dealt with this problem by predefining fixed niches using a Voronoi tessellation of the phenotypic space. Due to their fixed archive, both methods tend to reduce the variance of the solution set in the first iterations. Initial (random) samples tend to not cover the entire phenotypic space and thus competition is harsher, leading to the excluding of many solutions in the beginning. To maximize the number of available training samples for the VAE, the Voronoi-Elites (VE)~\cite{Hagg2020b} algorithm is therefore more appropriate.
VE does not precalculate the niches. It accepts all new artifacts until the maximum number of niches is surpassed.
Only then the pairs of elites that are phenotypically closest to each other are compared, rejecting the worst-performing pair members.

\begin{figure}[tb]
	\centering
	\includegraphics[width=\linewidth]{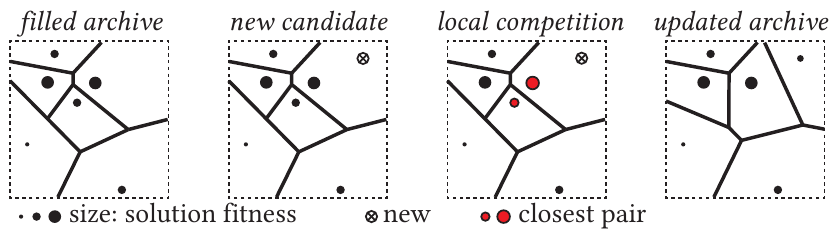}
	\caption{Updating a Voronoi archive. 
	A new candidate artifact is compared to the closest artifacts. The worse of the two is rejected or removed from the archive and the artifact with higher fitness is kept or added to the archive. Here the maximum number of niches is set to six and the borders between niches are drawn to illustrate each item's range of influence and how it changes after an update.}
	\label{img:VE}
\end{figure}

The VE archive's evolution is illustrated in Fig.~\ref{img:VE}. Selection pressure is applied based on artifact similarity. In effect, VE tries to minimize the variation of distances between artifacts in the (unbounded) archive. The total number of niches/artifacts is fixed, independent of the number of archive dimensions. Tournament selection is used to select artifacts from the archive. New artifacts are created by mutation, drawn from a normal distribution.

\section{Study Setup}

When a VAE is used for generation or search, the diversity of its output is bound by the expressivity of its latent space. The objective of our study is to analyze the generative capabilities of a VAE's latent space and give empirical evidence for its limitations. 

This section outlines the details of our study's subject domain, the generation of two-dimensional shapes, lists the general configurations of the VAE and VE algorithm (specific settings for experiments can be found in the experimental setups below) and explains how the two methods are combined to build two versions of a generative system which we compare in a series of experiments.

\subsection{Shape Generation}
\label{sec:domain}

\begin{figure}[bt]
    \centering
    \includegraphics[width=\linewidth]{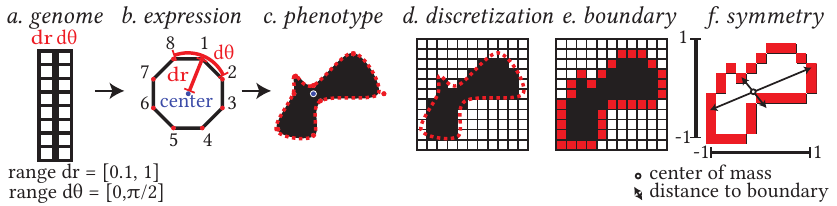}
    \caption{Generation of shapes: from (a) 16 genes to (b) eight control points, freely placed in two-dimensional space, to (c) a smooth interpolated spline and (d) a final bitmap rendering. Quality evaluation: (e) the boundary of the shape is determined and (f) the shape's symmetry is measured from its center of mass.}
    \label{fig:encoding}
\end{figure}

For our study, we focus on the generation of two-dimensional shapes, similar to a data set which has been proposed for the evaluation for the quality of disentangled representations~\cite{higgins2016beta}. Here we explain the setup of our shape generating system in the context of its later use with the VE algorithm.
The shapes are generated by connecting eight control points which can be freely placed in a two-dimensional space. Each control point is defined by two parameters, the radial ($dr$) and angular deviation ($d\theta$) from a central reference point (Fig.~\ref{fig:encoding}b). 
These 16 parameters serve as genomes, encoding the properties of each individual.
To form a final smooth outline, the points are connected by locally interpolating splines~\cite{catmull1974} (Fig.~\ref{fig:encoding}c). 
A discretization step renders this smooth shape onto a square grid resulting in a bitmap of $64\times64$~pixels (Fig.~\ref{fig:encoding}d). 

\subsection{Fitness}
\label{sec:fitness}

As a simple objective and fitness criterion, we have chosen point symmetry, which acts as an exemplary problem in generative applications, is easy to understand, and is computationally inexpensive.
To determine an artifact's quality, first, the boundary of the artifact is determined (Fig.~\ref{fig:encoding}e). Second, the coordinates of the boundary pixels are normalized to a range of -1 to 1 in order to remove any influence of the shape's scale (Fig.~\ref{fig:encoding}f). Third, the center of mass of the boundary is determined to serve as the center of point symmetry. Fourth, the distances to the center of pixels opposite of each other w.r.t. the center of mass are compared. Finally, the symmetry error $E_s$, the sum of Euclidean distances of all $n/2$ opposing sampling locations to the center, is calculated (Eq.~\ref{eq:fitness}). A maximally symmetric shape is one for which this sum equals zero. The fitness function $f_P(\boldsymbol{x})$ is calculated as follows:

\begin{equation}
f_P(\boldsymbol{x}) = \frac{1}{1 + E_s(\boldsymbol{x})} \hspace{2em} E_s(\boldsymbol{x}) = \sum_{j=1}^{n/2}\left\lVert \boldsymbol{x}_j-\boldsymbol{x}_{j+n/2}\right\rVert
\label{eq:fitness}
\end{equation}

\subsection{VAE Configuration}
\label{sec:VAEconfig}

\begin{figure*}[tbp]
	\centering
	\includegraphics{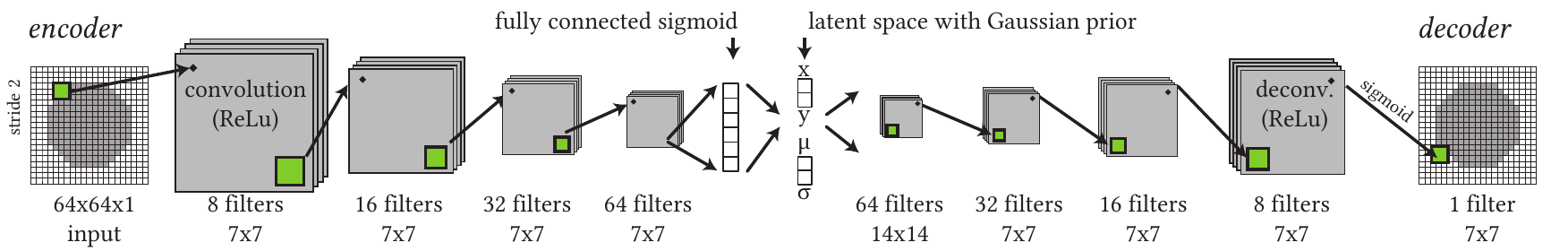}
	\caption{Architecture of a convolutional variational autoencoder.}
	\label{img:cvaeArchitecture}
\end{figure*}

Throughout this work, we use a VAE with a beta-annealing loss term~\cite{bowman2016generating,burgess2017understanding} and its decoder as a mapping network from latent codes to phenotype bitmaps (see Fig.~\ref{img:cvaeArchitecture}).
The model's encoder network is made up of four downscaling blocks, each consisting of a convolution layer (8, 16, 32 and 64 filters respectively; kernel size $7\times7$; stride 2) followed by a ReLU activation function. The set of blocks is followed by a final fully-connected layer. The decoder network inversely maps from the latent space to bitmaps through five transposed convolution layers, which have 64, 32, 16, 8 and 1 filter respectively, kernel size $7\times7$ and stride 2, except for the first layer which has a kernel size of $14\times14$. The last layer is responsible for outputting the correct size ($64\times64$ pixels). The weights of both networks are initialized with the Glorot initialization scheme \cite{glorot2010understanding}. 
The regularization term scaling factor $\beta$ was set to $4$ and the annealing factor $\gamma$ was increased from 0 to 5 over the course of the training, in order to focus on  improving the reconstruction error first, and improve the distribution in latent space later in the process. Each model was optimized with the Adam optimizer~\cite{kingma2015adam} with a learning rate $\mu = 0.001$ and a batch size of 128.

\subsection{VE Configuration}
\label{sec:VEconfig}

We configure VE to start with an initial set of samples, generated from a Sobol sequence~\cite{sobol1967distribution} in parameter space. Sobol sequences are quasi-random and space-filling. They decrease the variance in the experiments but ensure that the sampling is similar to a uniform random distribution and easily reproduced. In all experiments, VE runs for 1,024 generations, producing 32 children per generation. Children are produced by adding a small mutation vector, drawn from a normal distribution centered around zero with $\sigma=0.1$, to selected parent individuals. The selection is drawn at random from the archive. The number of artifacts in the archive remains constant, identical to the initial population size, over the entire experiment.

\subsection{Combining a VAE with VE into a Generative System}
\label{sec:combining}

\begin{figure}[tb]
    \centering
    \includegraphics[width=\linewidth]{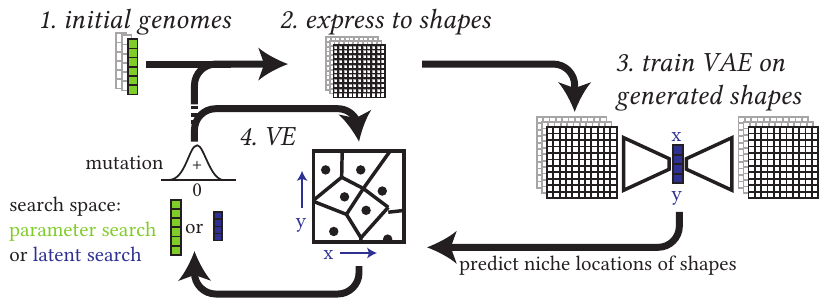}
    \caption{\textbf{AutoVE} combines a VAE and VE into a generative system in two phases. First, initialization: (1) a random set of genomes is generated and (2) converted into shape bitmaps which are used to (3) train a VAE. Second, optimization loop: (4) VE iteratively updates the archive of candidates. We compare two setups of this loop: the VE performs search either in parameter space or in the VAE's latent space.}
    \label{fig:generativeprocess}
\end{figure}

The VAE is combined with VE to form the AutoVE~\cite{Hagg2020b} generative system with the objective to produce point-symmetric two-dimensional shapes.
The difference to the original formulation is the use of a VAE instead of a classical autoencoder, as a VAE creates a more even occupancy of training samples in latent space as well as allowing interpolating new examples and disentangling latent dimensions.
The full generative process is illustrated in Fig.~\ref{fig:generativeprocess} and can be separated into two phases: 1) initialization and 2) an evolutionary optimization loop.
At initialization time, a set of random genomes is drawn and translated into bitmaps, their phenotypic counterpart. The VAE is trained to convergence on this set of bitmap data. The learned latent space is then used in the following evolutionary process and the model's encoder and decoder networks serve as mapping functions between the phenotypic bitmap representations and the model's latent representations and vice versa.

In the evolutionary optimization loop, the VE algorithm iteratively updates the archive and tries to increase the diversity as well as the quality of the archive through local competition. To compare two candidates to each other, it relies on the VAE's low-dimensional latent representations, which preserve semantically meaningful distances. We perform this optimization process in two different search spaces for the central comparison of our study: 1) parameter space (the explicit genome encoding) and 2) the VAE's latent space (the learned representation). In this way, we can evaluate the expressivity of a VAE's latent space and its capability to generate a diverse set of artifacts in comparison to the full space of possibilities which is reflected by the 16 predefined genetic parameters. The performance of the two approaches is measured in terms of diversity of the produced set (more on our diversity metric in the following Section~\ref{sec:diversitymetric}). This setup allows us to study the limitations of the latent space of a VAE and compare it to the baseline diversity of searching for candidate solutions over the possible parameters.

\subsection{Diversity Metric}
\label{sec:diversitymetric}
In the QD community, metrics that measure the diversity of a solution set are usually domain-dependent or require to take one of the QD algorithms as a baseline~\cite{Hagg2020b}. Archive-dependent metrics do not generalize well and introduce biases. We therefore only use distance-based diversity metrics that are calculated on the expressed shapes directly. Pure diversity (PD) measures diversity within a set of artifacts. We use the $L^{0.1}$-norm, which is suitable for high-dimensional cases~\cite{Wang2017b}, to find the minimum dissimilarity between an individual item and the items in a set $X$ (Eq.~\ref{eq:distance}). The PD value of a set $X$ is calculated recursively and is equal to the maximum of the sum of its value on all but one of the members and the minimum distance of that member to the set (Eq.~\ref{eq:PD}). 

\begin{equation}
d(s,X) = \underset{s_i \in X}{\mathrm{min}}(L^{0.1}(s,s_i))
\label{eq:distance}
\end{equation}

\begin{equation}
PD(X) = \underset{s_i \in X}{\mathrm{max}}(PD(X-s_i)+d(s_i,X-s_i))
\label{eq:PD}
\end{equation}

\noindent
PD was first proposed in the context of many-objective optimization~\cite{Wang2017b} and has been applied to high-dimensional phenotypes~\cite{Hagg2020b}. PD can deal with a high number of dimensions and is consistent with some other widely used diversity metrics. By calculating PD on a set of bitmaps, we can measure diversity directly, independent of the representation in parameter space or the VAE's latent space.

\section{Experiments}
\label{sec:experiments}

It is commonly assumed that GM, such as a VAE, have good interpolative and reasonable extrapolative capabilities, which makes their latent space a potentially appealing search space. But how well a search in this space performs in terms of generating a diverse output, to our knowledge, has not yet been adequately investigated. In the setup of our generative system\footnote{The code to reproduce the experiments can be found at \url{https://github.com/alexander-hagg/ExpressivityGECCO2021}.} the latent space of a VAE is used to search for and generate two-dimensional shapes in the form of square bitmaps. We compare the output diversity of this process to the baseline diversity of a search performed on the explicit genome encoding (parameter space).
We aim to gain insight into two questions:
1) how accurately can a VAE represent a variety of shapes, that is to say how useful are its latent representations, and 
2) how well can a VAE generate unknown shapes?

\begin{figure}[bt]
    \centering
    \includegraphics[width=\linewidth]{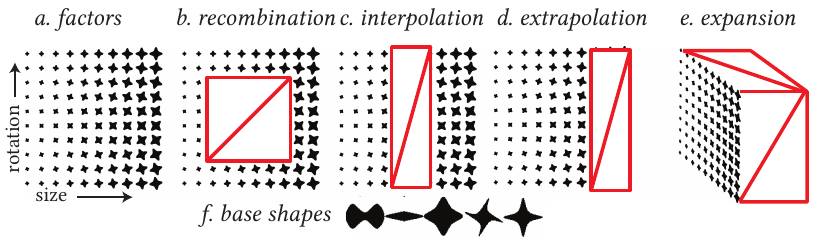}
    \caption{(a) generative factors used to create data sets in this work; (b--e) four tasks on which we compare the performance of latent space search with parameter search, the red rectangles indicate artifacts that either have been left out of a data set (b, c, d) or are not available (e); (f) all base shapes used in this work.}
    \label{fig:tasks}
\end{figure}

All data sets in our first experiment consist of samples which have been produced by varying two generating factors: scale and rotation (Fig.~\ref{fig:tasks}). We present here a series of corresponding tasks that we evaluate in two experiments:
\begin{enumerate}[label=\alph*)]
    \item With a complete set of samples as a baseline data set we evaluate the standard reconstruction error of the model in order to determine the general quality of latent representations.
    \item In the recombination task, we leave out a subset of artifacts in the center of the ranges of values of both generating factors, leaving sufficient examples at either end of the ranges.
    \item In the interpolation task, the left-out subset of artifacts covers the complete range of one of the two generating factors, while for the other some examples remain at both ends of its range of values.
    \item The extrapolation task consists in omitted samples at one end of values of one factor of variation, which affects the complete range of the other factor.
    \item The expansion tasks focuses on generating artifacts beyond the two given generating factors from the complete data set.
\end{enumerate}

The VAE is expected to perform reasonably well in recombining (b), interpolating between (c) and extrapolating beyond the available variations (d) to reproduce the samples missing from the training data. In the expansion task (e), we expect the VAE's latent space to only produce artifacts of poor quality outside of the generating factors present in the training data.

\begin{figure*}[hbt]
    \centering
    \includegraphics[width=\linewidth]{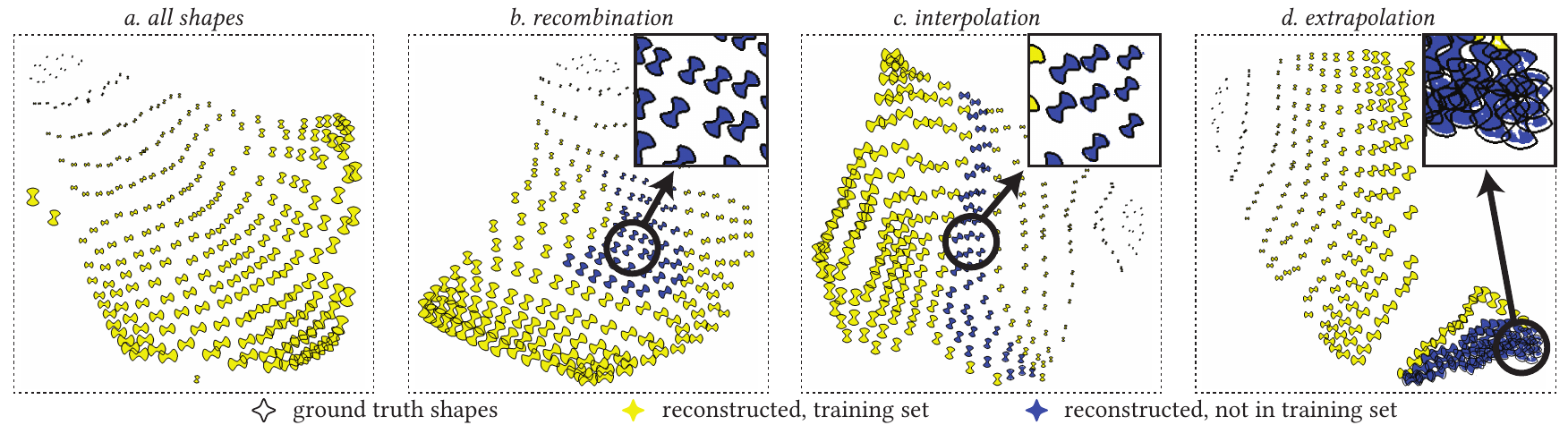}
    \caption{Examples of samples and latent spaces produced by the VAE with a latent dimensionality of eight (projected to two dimensions by t-SNE). Shapes in yellow represent samples that were present in the data set, while blue ones were not and have been generated by the model. Black outlines show the ground truth shapes, the difference between the outlines and shapes accounts for errors in reconstruction.}
    \label{fig:generatedshapes}
\end{figure*}

\subsection{Recombination, Interpolation and Extrapolation}
\label{experiments:I}

We train one baseline VAE on the complete set of variations of a base shape (Fig.~\ref{fig:tasks}a,f) (256 shapes, scaled by factors of 0.1 to 1.0 and rotated by 0 to ${\frac{\pi}{2}}$ in 16 steps each) and three additional models, each one on the data set of one special task (b--d) with held-out samples. The VAEs are trained for 3,000 epochs, after which we choose the models with the lowest validation error (calculated on 10\% of the input data).

To determine whether the VAE can correctly reproduce, and thus properly represent, the given shape, we measure the models' reconstruction errors. For the baseline model this is done over the complete data set. For the task models (b--d) the reconstruction error is calculated only on the held-out examples. We define the reconstruction error as the Hamming distance between an input bitmap and a generated bitmap, normalized by the total number of pixels. The Hamming distance is useful to measure differences between bitmaps, due to their high dimensionality. A high reconstruction error would indicate that the model cannot properly generate the shapes and that its latent space does not provide an adequate search space for VE. Generating shapes to which there are no corresponding training examples, the reconstruction errors of unseen shapes that can be created with recombination and interpolation (b and c) are expected to be lower than for extrapolation (d).

To determine the resolution of the models, we measure the distances in the latent space between the training examples for the baseline model and between the training and the unseen examples for the task models (b--d). If the latter are of a similar order of magnitude as the first, the models are able to distinguish unseen shapes from the training examples and from each other. This would indicate that the model's resolution is high enough to provide features of sufficient quality to perform a VE search.

This experiment was performed separately on each of the five base shapes (Fig.~\ref{fig:tasks}f) and for three different sizes of the VAE's latent space (4, 8, and 16 dimensions), as we assumed that the model would not able to perfectly learn the two generating factors. The results are reported as averages over the resulting 15 total runs.

\paragraph{Results}

\begin{figure}[tb]
    \centering
    \includegraphics[width=\linewidth]{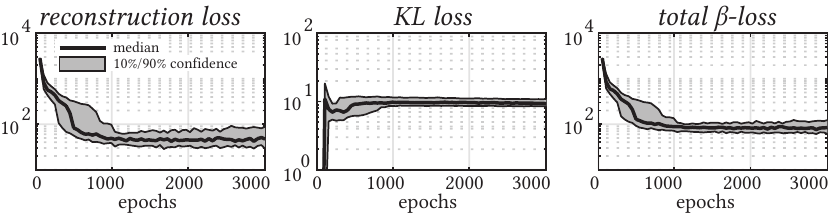}
    \caption{VAE validation losses during training.}
    \label{fig:lossesI}
\end{figure}

Fig.~\ref{fig:lossesI} shows the reconstruction, KL and total $\beta$-loss on the validation data, during training of the models. The training does not need much more than 1,000 epochs to converge.

\begin{figure}[tb]
    \centering
    \includegraphics[width=\linewidth]{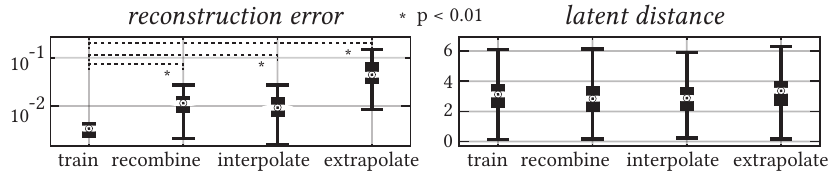}
    \caption{Reconstruction errors and latent distances for tasks a-d.}
    \label{fig:errorsAndDistances}
\end{figure}

Fig.~\ref{fig:errorsAndDistances} (left) shows the reconstruction errors for the training, recombination, interpolation and extrapolation sets. The error the models produce on the training samples is lower than when reproducing the recombination and interpolation sets. As expected, the error on the extrapolated shapes is highest. All significant differences (two-sample t-test, $p < 0.01$) between the reconstruction error on the whole data set and a hold-out set are marked with an asterisk. The latent distances between the shapes in the four sets are shown in Fig.~\ref{fig:errorsAndDistances} (right). The distance distributions are similar.

Four exemplary latent spaces are shown in Fig.~\ref{fig:generatedshapes} from models with a latent space dimensionality of 8, which has been projected to two dimensions with the dimensionality reduction method t-distributed stochastic neighbourhood embedding (t-SNE)~\cite{VanDerMaaten2008}. The first latent space (a) corresponds to the baseline model, trained on the complete training set. The other visualizations (b, c, and d) show the three tasks in which some shapes have been omitted.

\subsection{Expansion}
\label{experiments:II}

The last task, expansion (e), cannot be treated as per the previous experiment, because we cannot easily define an a priori ground truth shape set ``outside of latent space''. Instead, we compare the two search spaces (parameter: PS, and latent: LS) using the framework proposed in Fig.~\ref{fig:generativeprocess}. We measure which one of the two search spaces produces the most diverse set of artifacts using the PD metric explained above. The experiment is split up into two configurations. 

In the first configuration (\textbf{R}), both of the compared search approaches start from the same random initial set of genomes, which is common in many optimization problems. We increased the size of the set to 512, as this experiment poses a more difficult optimization problem. The genomes are translated into bitmaps, which serve as the training data for a VAE model. VE is then performed in both search spaces to fill two separate archives of 512 shapes each. The resulting shape sets are compared w.r.t. their diversity and average fitness, which are often in conflict with each other. As the translation from genome to bitmap always produces a contiguous shape, it is reasonable to expect that a VAE would learn to produce shapes, and not only random noise, even when starting with a randomly generated set of examples. 

Often a generative system does not start from a random set of data, but rather a set of examples that has been observed in the real world. A second configuration, continuation (\textbf{C}), is defined to reflect this. We ask whether the diversity improves when training a VAE with a set of high-quality generated artifacts from a previous VE search.
We use the archive of shapes produced by PS from the random initial set (\textbf{R}) as training data for a new VAE model. Both PS and LS are then performed again with this improved model.

It is expected that LS will interpolate between training samples, but not be able to expand beyond the generative factors in the data, except through modeling errors. Since PS is performed in the encoding's parameter space, this search approach should be able to produce a higher artifact diversity in both configurations \textbf{R} and \textbf{C}.

The number of latent dimensions of the VAE has been set to 8, 16 and 32 to analyze the influence of the degrees of freedom in latent space, when it is lower than, equal to, or higher than the number of parameters of the genome representation. A higher number of degrees of freedom gives an advantage to the latent model, a lower number would give it a disadvantage. When using 16 latent dimensions, VE deals with the same dimensionality in PS and LS. The number of filters in the VAE is quadrupled to give the model a better chance at learning the larger number of variations.

This experiment has been repeated 10 times per configuration: 1) random initial set \textbf{R} in PS, 2) continuation \textbf{C} in PS, 3) \textbf{R} in LS and 4) \textbf{C} in LS.

\begin{figure}[bt]
    \centering
    \includegraphics[width=\linewidth]{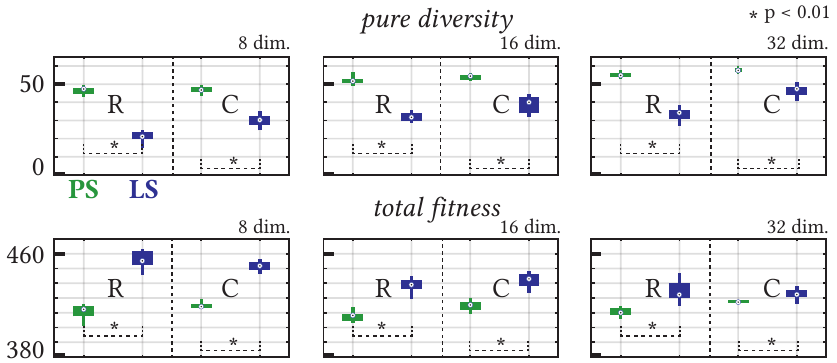}
    \caption{Diversity (top) and total sum of fitness (bottom) of artifact sets of both parameter (PS, green) as well as latent search (LS, blue). VAEs were trained with 8, 16 and 32 latent dimensions respectively. Both the random initialization (\textbf{R}) and continuation (\textbf{C}) configurations of the experiments are shown (in every box the two left-hand bars correspond to \textbf{R} and the two right-hand to \textbf{C}). Significant differences (two-sample t-test, $p < 0.01$) are marked with an asterisk.}
    \label{fig:pdfit}
\end{figure}

\paragraph{Results}

\begin{figure}[tb]
    \centering
    \includegraphics[width=\linewidth]{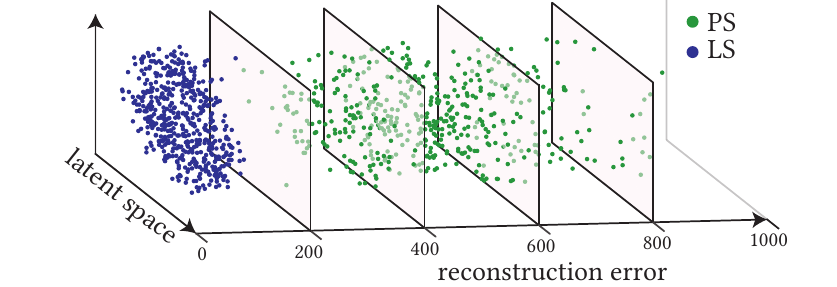}
    \caption{Expansion in a 16-dimensional latent model (projected to two dimensions with t-SNE). We interpret the reconstruction error of a shape as its distance from the latent surface. Samples from parameter search (PS, green) tend to extrapolate away from the latent distribution (LS, blue).}
    \label{fig:expansion}
\end{figure}

Fig.~\ref{fig:pdfit} compares the PD and total fitness of the generated artifact sets. The diversity of PS is significantly higher than LS. In turn, LS produces artifacts with higher levels of fitness. Although the difference between PS and LS gets smaller when continuing search from an updated model (configuration \textbf{C}), it is still significant. Again, all significant differences (two-sample t-test, $p < 0.01$) between random initialization and continuation in all configurations are marked with an asterisk. Fig.~\ref{fig:expansion} shows the expansion away from the latent surface achieved by PS, analogous to our previous hypothesis (Fig.~\ref{fig:tasks}e). For this visualization, the PS and LS artifacts' position in the 16-dimensional latent space is reduced to two dimensions using t-SNE. The reconstruction error of the model's prediction of the PS artifacts is used as a distance measure to the latent surface. 
An example of the resulting shape sets of PS and LS is shown in Fig.~\ref{fig:QDshapes} to illustrate the effective difference in pure diversity.

\section{Discussion}

VAEs are able to produce previously unseen examples through recombination and interpolation as expected (Section~\ref{experiments:I}). The more difficult task, extrapolation beyond the extremes of the generative factors, results in a higher reconstruction error (Fig.~\ref{fig:errorsAndDistances}). The distributions of latent distances between all four data set variants are similar. This suggests that, even when VAEs are not able to reproduce the extrapolated shapes, they can still distinguish them from the training data and from each other. That is to say, the position of examples in the latent space reflects their semantic relationship, as visualized in Fig.~\ref{fig:generatedshapes}. Shapes are not properly reconstructed in the extrapolation task, but they are still positioned in a well-structured relation to others. The presented evidence leads to the hypothesis that expansion away from the latent surface is more difficult when searching the latent space directly.

We further examine the results of the expansion task (Section~\ref{experiments:II}).
The ability of a VAE to find new shapes is indirectly measured by comparing the diversity of the artifact sets created by a parameter (PS) and a latent search (LS). The diversity of PS is significantly higher than that of LS, as is shown in Fig.~\ref{fig:pdfit}. This holds as the number of latent dimensions is increased beyond the number of degrees of freedom in the original encoding, or when the VAE is updated after a first VE run (\textbf{C}). Although a trade-off between diversity and fitness is expected, it becomes less pronounced in the 32-dimensional model. This provides evidence for the conclusion that PS indeed finds a more diverse set of artifacts than LS. 
We therefore recommend to use a more expressive predefined parametric encoding, whenever it is available, rather than the extracted feature space of a GM such as a VAE. Yet, a GM's latent space is still useful for its ability to distinguish shapes and its semantically meaningful mapping to lower-dimensional space.

\section{Conclusions}

In this work, we have presented a systematic study on the limitations of latent spaces of deep GMs as a base for divergent search methods, specifically the VE algorithm. Our findings quantify a VAE's ability to generate samples through recombination, interpolation and extrapolation within and expansion beyond the distribution of a given data set. We compare the diversity of generated artifacts when VE is run either in latent space or parameter space. Our findings show that the pure diversity of artifact sets generated by latent space search is significantly lower than that of parameter space search. Based on these observations we recommend using a VAE's latent space as an approximate measure of similarity. Evolutionary search for the generation of diverse outputs should, however, preferably be performed on explicitly expressed genome parameters, whenever these are available. The expressivity of a VAE, when used as a generator for diverse artifact sets, is limited by the generative factors in the training data.

Our findings are limited to multimodal continuous domain optimization.
The presented conclusions are only meaningful to problem settings which allow for multiple solutions. Neither did we extend them to combinatorial search.
The presented problem setting is kept simple in order to remain illustrative and easy to interpret. Most application domains are much more complex and more work is required to confirm or refute our assumptions on generalization.

We plan to extend these first results with a systematic study of the individual parts of our setup and their influence on expressivity. 
We will look at different priors for a VAE latent distribution, 
the size of the training data set, and mapping to a higher-dimensional latent space.
Another opportunity for further study is the architecture of the generative model. Comparing the performance of a VAE to that of an autoregressive or flow-based model, a GAN or a transformer could highlight strengths or weaknesses of individual modelling methods, and allow for a more general understanding of their generative capabilities.

The usefulness of a GM's ability to interpolate, extrapolate or expand has to be discussed in the context of its application. In one setting, it might be ideal to perfectly reproduce a given domain and a model might be considered as working well if it can fit a distribution accordingly. In another context, however, and here we do include applications with a focus on exploration of possibilities and discovery, a systems's ability to surprise through its unexpected outputs can be of value and a desirable quality.

In the real world, the assumption that a GM can learn a ``perfect'' representation does not hold. Latent search spaces are not guaranteed to cover the whole possibility space as defined by the underlying parameters of the system that produced the training examples. The idea that a GM can fit the ``true'' distribution overlooks the fact that the data might not cover the entire range of parameters. Furthermore, a perfect model does not produce anything unexpected, only creating high quality artifacts with low diversity. A broken model, on the other hand, might not produce anything useful. 
The use of modeling errors to find novel artifacts is certainly a mechanism that can allow us to find novel solutions within the model. An important question for future work is whether we can use early stopping when training models to create novel yet useful artifacts: is there a correlation between training loss and diversity?

We can assume that GMs are always limited by the data we give them, which biases models towards the most prominent features therein. Yet, in very high-dimensional domains for which we can collect large data sets, like image and video data, a search in latent space already affords a vast amount of possible outcomes, which might be sufficient for some applications. 

Surprisingly, using generative models to understand diversity and guide evolutionary computation produces more diverse sets of artifacts than having the models generate the artifacts themselves. With this work, we hope to have contributed some inspiration to using generative models in novel ways.

\begin{acks}
The authors thank the anonymous reviewers for their valuable comments and helpful suggestions. S. Berns is funded by the \grantsponsor{IGGI}{{EPSRC} Centre for Doctoral Training in Intelligent Games \& Games Intelligence ({IGGI})}{https://iggi.org.uk} [\grantnum{IGGI}{EP/S022325/1}].
\end{acks}

\bibliographystyle{ACM-Reference-Format}
\bibliography{bibliography}

\end{document}